\documentclass[10pt,twocolumn,letterpaper]{article}

\usepackage{iccv}
\usepackage{times}
\usepackage{epsfig}
\usepackage{graphicx}
\usepackage{amsmath}
\usepackage{amssymb}
\usepackage{multirow}

\usepackage[pagebackref=true,breaklinks=true,letterpaper=true,colorlinks,bookmarks=false]{hyperref}

\iccvfinalcopy 

\def\iccvPaperID{10} 

\ificcvfinal\fi
\begin{document}

\title {Residual Attention Graph Convolutional Network for Geometric 3D Scene Classification}

\author{Albert Mosella-Montoro and Javier Ruiz-Hidalgo\\
Universitat Polit\`ecnica de Catalunya\\
Image Processing Group - Signal Theory and Communications\\
{\tt\small https://imatge.upc.edu} \\
{\tt\small \{albert.mosella, j.ruiz\}@upc.edu}
}

\maketitle
\ificcvfinal\thispagestyle{empty}\fi

\begin{abstract}
Geometric 3D scene classification is a very challenging task. Current methodologies extract the geometric information using only a depth channel provided by an RGB-D sensor. These kinds of methodologies introduce possible errors due to missing local geometric context in the depth channel. This work proposes a novel Residual Attention Graph Convolutional Network that exploits the intrinsic geometric context inside a 3D space without using any kind of point features, allowing the use of organized or unorganized 3D data. Experiments are done in NYU Depth v1 and SUN-RGBD datasets to study the different configurations and to demonstrate the effectiveness of the proposed method. Experimental results show that the proposed method outperforms current state-of-the-art in geometric 3D scene classification tasks. 
\end{abstract}

\section{Introduction} \label{sec:intro}

Scene classification is a fundamental problem in computer vision. It has a wide range of practical applications such as semantic recognition~\cite{semanticRecApp} and remote sensing \cite{remoteSensingSceneClassification} among others. This paper focuses on indoor scene classification. Indoor scene classification is a challenging task due to the fact that the same class of scene can have large variations in light, shapes, occlusions and different layouts. Some of these challenges have been proved very difficult to solve without the 3D information that is lost in image capturing. 

During the last years, the use of sensors able to capture 3D data has dramatically increased. These sensors are able to capture information directly into a 3D point cloud representation ({LiDAR}) as well as capturing registered colour and depth information (Kinect). One of the challenges that are inherent in some of these sensors is the fact that the captured 3D point clouds are unorganized. This increase in usage has introduced the necessity to develop new methods to understand this data.

Convolutional Neural Networks (CNNs) have achieved extremely good performance in a multitude of tasks, such as in image classification~\cite{VGG,ResNet} or segmentation~\cite{FCNSemanticSegmentation, pspNet}. However, standard CNNs can not be used to process 3D point clouds directly. Standard convolution operations only work in a lattice structure in the euclidean space which is not the case for 3D point clouds.

This paper proposes a new methodology that exploits the geometric information of 3D point clouds using graph neural networks. These 3D point clouds can be obtained from an RGB-D image or directly from {LiDAR} sensors. The main contributions of this paper are: a) The use of an Attention Graph Neural Network to capture the geometric information of a scene. b) The adaptation of Attention Graph Neural Networks into a deep residual architecture similar to~\cite{ResNet,GatedGraphConv}. c) The proposed network has been applied into the 3D geometric scene classification task on two datasets (NYU Depth v1 and SUN-RGBD) outperforming current state-of-the-art on this task.

\section{Related work} \label{sec:rw}

\subsection{Geometric Learning}\label{sb:gl}

Currently, there are different techniques to process 3D point clouds using geometric information. Multi-view based techniques~\cite{MultiViewSu3DShapeRecog, SnapNet, SnapNetR, dai3DMV} represent a 3D space as a collection of 2D views where standard {CNNs} are used. However, the geometric information used in multi-view based techniques is quite limited. To overcome this limitation, the use of voxel grids has been proposed~\cite{VoxNet, 3DShapeNets, VolumetricMultiViewCNN, ScanNet, SEGCloud}. Working with voxel grids can be inefficient due to the cubic complexity associated with this kind of structure. Related to voxel grids, researchers have suggested using hierarchical spatial data structures, for example, kd-trees and octrees, which are more memory and computation efficient~\cite{OctreeGeneratingNetworks}. Another approach to tackle these kinds of data is by applying template on a neighborhood representation witch is obtained by mapping the neighbors on a fixed structure, as it is done in MoNet~\cite{MoNet}.

This work focuses on the representation of 3D point clouds as a graph. Currently, there are two main ways to use graph data on neural networks. \emph{Graph Neural Networks}~\cite{ScarselliGraphNeuralNetwork}~\cite{Qui3DGNNSS}, which recurrently apply neural networks to every node of the graph, and \emph{Graph Convolutional Networks}~\cite{BrunaSpectralNetworks}, which use a generalization of the convolution to a graph. This generalization can be done on the \emph{spectral domain} and the \emph{spatial domain}.

On the one hand, \emph{spectral domain} makes use of the graph spectral analysis theory, where the convolution corresponds to the multiplication of the signal on vertices transformed into the spectral domain, using the graph Fourier transform. The spatial locality of filters is given by the smoothness of the spectral filters, in the case of~\cite{BrunaSpectralNetworks}, modeled as B-splines. This spectral transformation implies multiplications with the eigenvector matrix which have a high computational cost. \emph{Defferrard et al.}~\cite{DefferrardCNNGSpectral} propose a parameterization of filters as Chebyshev polynomials that is computationally more efficient. In all cases, methods that use spectral domain have the drawback that the graph structure (such as the number of nodes) must be fixed.

On the other hand, \emph{spatial domain} methods~\cite{DuvenaudCNG, kipf2017semi, peernets} define convolutions directly on the graph, operating on groups of spatially close neighbours. Furthermore, it is proposed to add the signals provided by each node in the neighbourhood and do multiplication using a weight matrix that shares the weights between all edges. \emph{Edge-Conditioned Convolutions}~\cite{SimonovskyECC} proposed by \emph{M. Simonovsky and N. Komodakis}, performs convolutions over local graph neighbourhoods exploiting the attributes of the edges. An intuitive explanation of the proposal is that the lattice space that is needed to do a convolution is artificially created using the edges. These edges have a direct influence on the weights of the filter used to calculate the convolution. \emph{Verma et al.}~\cite{FeaStNetCVPR2018} propose \emph{FeaStNet}.This operator tries to establish correspondences between filter weights and graph neighbourhood. These correspondences are dynamically computed from the node features using a linear layer. A recent work by \emph{Wang et al.}~\cite{dgcnn} proposes a Dynamic Edge Convolution operation, that computes the new feature of each node using an aggregation operation over the resulting values obtained by a  multi-layer perceptron (MLP) applied on each neighbourhood.

The proposed work in this paper is based on \emph{Edge-Conditioned Convolutions} in order to create the Graph Convolutional Network to process 3D point clouds. It is also inspired by the investigation done by \emph{X.Bresson et al.}~\cite{GatedGraphConv} that demonstrates the benefits of adding residual learning on Graphs Convolutional Networks. The proposed extension adapts Edge-Conditioned Convolutions to work in a deep residual learning architecture. 

\subsection{Attention Methodologies}

Visual attention enables humans to analyze complex scenes and devote their limited perceptual and cognitive resources to the most important of sensory data. Attention models aim to automatically identify the most attractive regions in images like the human visual systems do. \emph{Xu et al.}~\cite{visualAttentionNeuralImage} introduced an attention based model that automatically learns to describe the content of images. Their model is able to automatically learn to fix its gaze on salient objects. \emph{Ren et al.}~\cite{instanceSegmentationRecurrentAttention} propose an end-to-end Recurrent Neural Network architecture with an attention mechanism that produces detailed instance segmentation. Moreover, attention mechanisms are common on the machine translation field~\cite{machineTranslationAttention}. \emph{Velickovic et al.}~\cite{GAT2018} introduce an attention mechanism on graphs using the node information of a graph. The proposed attention network in this work is based on the intrinsic behaviour of the Edge Conditioned Convolutions~\cite{SimonovskyECC} and it is presented in subsection~\ref{subsec:AGC}.

\subsection{Scene Classification}
Traditional methodologies for scene classification make use of handcrafted features such as SIFT~\cite{SIFT} and HOG~\cite{HoG}. But with the emergence of deep learning techniques, better features can be obtained. Places-CNN~\cite{placesCNN} is the most successfully deep feature learning model in scene classification. Their work consist of training standard architectures, such as VGG~\cite{VGG} or ResNet~\cite{ResNet} using a very large provided dataset. Furthermore, \emph{George et al.}~\cite{semantincSceneRecognition} propose to model the occurrence patterns of objects in scenes, capturing the informativeness and discriminability of each object for each scene. \emph{Cai et al.}~\cite{rgbdSceneCategorizationMultimodal} propose a new CNNs multi-modal feature learning framework for RGB-D scene classification. This method can capture the local structure from the RGB-D scene images and automatically learn a fusion strategy. However, this method uses a 2D CNN with the depth channel as input. Possible errors can appear from missing part of local context information since the depth channel does not contain intrinsic parameters of the camera. In contrast, the method proposed in this paper obtains the geometric information taking advantage of the local context intrinsic in the 3D space.

\section{Residual Attention Graph Convolutional Networks} \label{sec:method}
In this section, an extension of Graph Convolutional Networks is presented to tackle the problem of 3D geometric scene classification. The extension is composed of four main parts: Graph construction~\ref{subsec:gc}, Attention Graph Convolution~\ref{subsec:AGC}, Residual Graph Convolutional block~\ref{subsec:RAGC} and the adaptation of the pooling mechanism to work with graphs~\ref{subsec:PGO}.

\subsection{Graph Construction}\label{subsec:gc}
Graph Construction is an important step on Graph Convolutional Networks as connections between nodes (edges) act as the receptive field on conventional CNNs. Edges indicate the influence between nodes in the graph. Graph Construction can be seen as three different stages: a) Project RGB-D image to 3D space. If the input is a 3D point cloud, this step can be skipped. b) Create the connectivity between nodes. Two methods will be explored: Radius proximity connection and $K$ nearest neighbours (kNN). Both have the particularity that the edges are directed. c) Add attributes to each edge of the graph.

Let $[x, y, z]$ be the 3D coordinates of a point in the camera coordinate system and $[u, v]$ the coordinates of a point in the image. The focal length of the camera is represented by $[f_x, f_y]$ and the coordinates of the principal point are $[c_x, c_y]$. The Pinhole camera model is used to project RGB-D image to 3D space as is described in Eq.~\eqref{eq:pinhole_projection}.

\begin{equation} \label{eq:pinhole_projection}
\left.
\begin{aligned}
    z &= depth channel\\
    x &= \frac{(u - c_x) \cdot z}{f_x}\\
    y &= \frac{(v - c_y) \cdot z}{f_y}
\end{aligned}
\right\}
\end{equation} 

Each projected point in the 3D space will be a node in the graph. In order to generate the edges, two different strategies will be explored. The first one is $K$ nearest neighbours ({kNN}). This approach consists of selecting the $K$ closest neighbours in the geometric space of each node of the graph and connect them. $K$ will be constant for all the nodes of the graph. This means that each node of the graph will have the same number of neighbours. The second one is radius proximity connection which consists of selecting all nodes in the graph that are inside of a defined radius $r_g$ as neighbours for each node. The radius will be constant for each node, therefore, each node may have a different number of neighbours. Both strategies need to do a self-connection to guarantee that their own node has influence from itself in successive layers of the network. 

Finally, attributes for edges and features for nodes need to be added. In this work the nodes of the input graph will not have any features and edges will have geometric attributes. Specifically, offsets between the two nodes connected. However, adapting node features and edge attributes to other domains is straightforward, giving the proposed network a great potential to adapt to different sensors (such as {LiDAR}).

\subsection{Attention Graph Convolution}\label{subsec:AGC}

\begin{figure*}
    \begin{center}
    \includegraphics[width=0.7\textwidth]{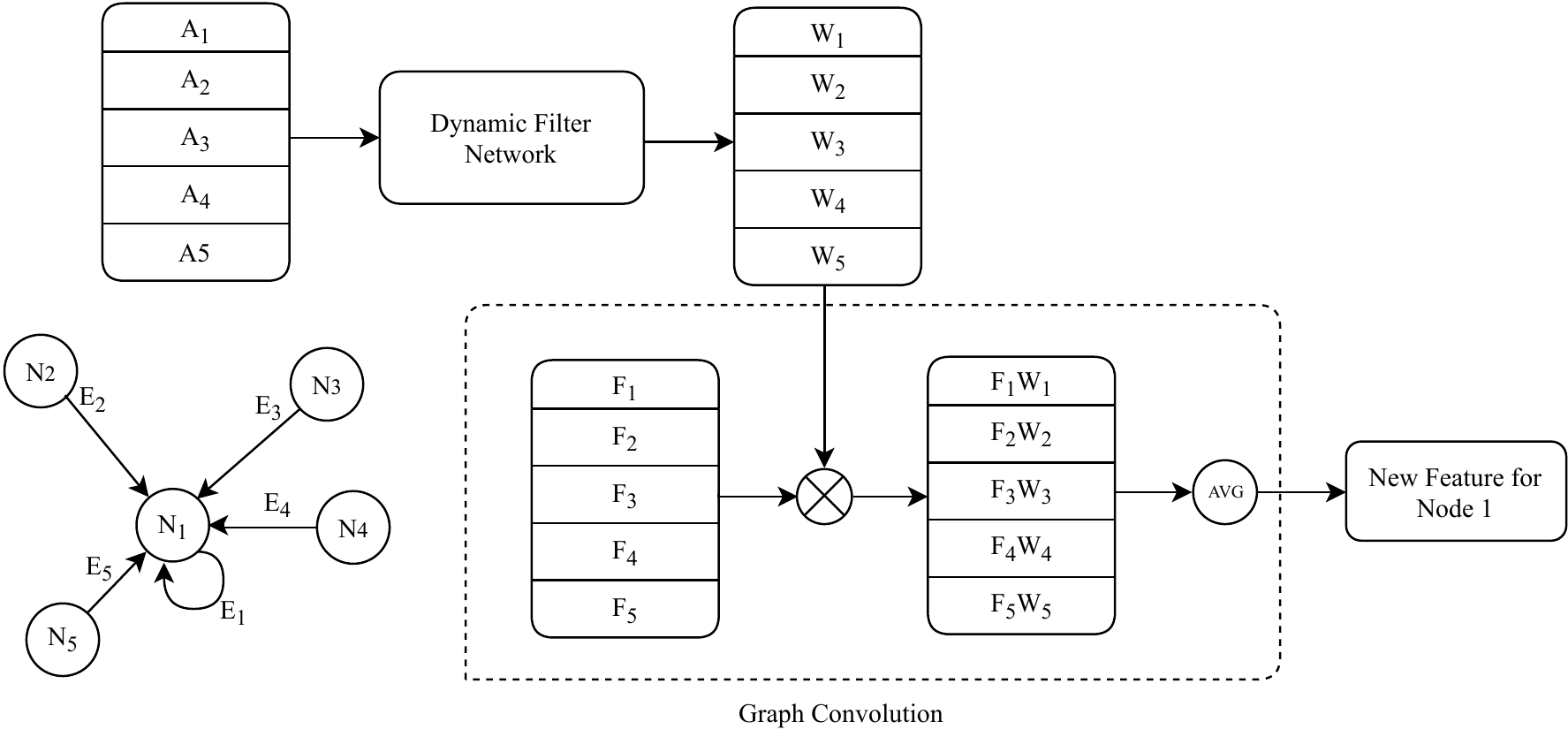}
    \caption{Picture of an AGC layer on a directed graph. Where $N$ is a node, $E$ an edge, $A$ the features of an edge, $F$ the features of a node and $W$ the weights of the filter.}
    \label{fig:ecc}
    \end{center}
\end{figure*}

Attention Graph Convolution (AGC) is an operation based on Edge-Conditioned Convolution proposed by \emph{Martin Simonovsky et al.}~\cite{SimonovskyECC}. This operation performs convolutions over local graph neighbourhoods exploiting the attributes of the edges. An intuitive explanation of the proposal is that the lattice space that is needed to do a convolution is artificially created using edges. These edges have a direct influence on the weights of the filter used to calculate the convolution. Depending on the edge attribute a weight will be generated. This generation of weights is done by a \emph{Dynamic Filter Network}~\cite{Dynamicfilter} which can be implemented with any differentiable architecture. In this work, it will be implemented using FC($x$) layers, where FC is a fully connected layer and $x$ the number of output features of the layers. This filter will be applied to each edge independently. The output of it will be ($d^l \cdot d^{l-1}$), where $d^l$ is the number of node features of layer $l$ and $d^{l-1}$ the number of node features of the previous layer. This is needed because this filter will generate the weights of the bank filters used on the convolution. Figure~\ref{fig:ecc} depicts the AGC layer operating over a node $N_1$ of an input graph. 
In section~\ref{subsec:ablation_studies}, different configurations of these \emph{Dynamic Filters} will be studied.

This convolution operation is formalized in Eq.~\eqref{eq:ecc} where $X$ is the vector of node features, $N$ the set of neighbourhoods, $\Theta$ is a matrix with the weights generated by the Dynamic Filter Network and $b$ a learnable bias of the layer. Index $i$ indicates the current node to evaluate, $l$ corresponds to a layer index in a feed-forward neural network and $j$ the neighbours of the node $i$. 

\begin{equation}
    X_i^l = \frac{1}{|N|} \sum_{j\in N(i)} \Theta^l_{ji} X_j^{l-1} + b^l
    \label{eq:ecc}
\end{equation}

As can be observed, the connectivity edge policy has a direct impact on the convolution operation. These connections can be seen as the receptive field of the graph convolution operation. If {radius} policy is applied, the receptive field of the convolution operation is equal for each node of the graph, like convolutions applied in the euclidean space. In contrast, if the {kNN} policy is chosen each node can have a completely different receptive field. It depends on the point cloud density. 

The attention stage of this operation is done in the weights generation step. The Dynamic Filter Network generates weights conditioned by the edge attributes of the neighbourhood. In the case of this work, these edges will have geometric attributes, that means, Dynamic Filter Network will pay attention to the nodes depending on their geometric information. 

\subsection{Residual Attention Graph Convolution}\label{subsec:RAGC}

\begin{figure}[htb]
    \centering
    \includegraphics[clip, trim=2.25cm 0cm 2.25cm 0cm, width=0.25\textwidth]{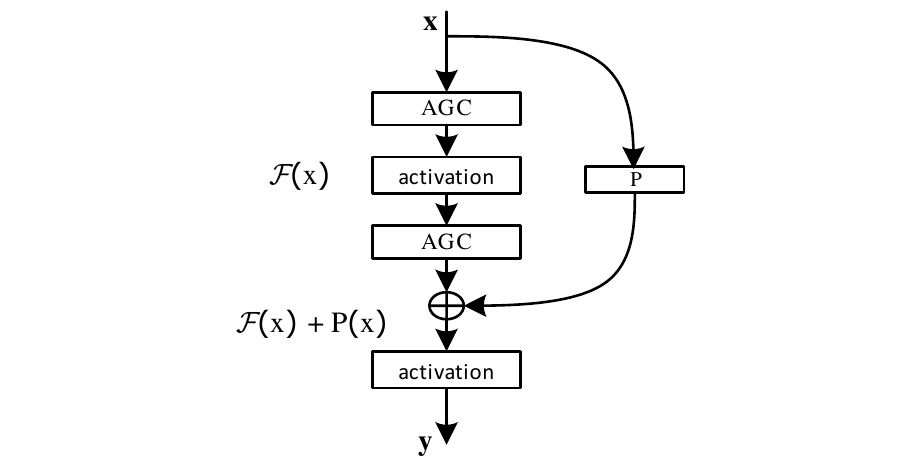}
    \caption{Schema of the Residual Attention Graph Convolution block.}
    \label{fig:ragc}
\end{figure}

The previous \emph{Attention Graph Convolution (AGC)} is extended to a \emph{Residual Attention Graph Convolution (RAGC)} following the inspiration of the ResNet~\cite{ResNet} architecture. \emph{Bresson et al.} demonstrated in~\cite{GatedGraphConv} that standard Graph Convolutional Networks can benefit from a residual learning approach. The proposal is to extend AGC layers with two stacked layers (more stacked layers could be employed). Figure~\ref{fig:ragc} shows a schema of the proposed RAGC block.

Eq.~\eqref{eq:recc} formalizes the residual learning extension, where $\mathcal{F}$ is a stack of AGC layers (including their corresponding activation map), $P$ a projection function, that projects the input to the same feature space that the output of the last AGC layer of the stack, $x$ and $y$ are the input and output of the stacked layers considered.

The projection function of the residual block, $P(x)$, is implemented as 1D convolution and it introduces extra parameters and computation complexity. It can be avoided, only if the dimensionality of the input and the output of the stacked layers are equal. However, keeping the projection function improves the performance of the network~\cite{ResNet}. 

\begin{equation}
    y = \mathcal{F}(x, \{AGC_i\} + P(x))
    \label{eq:recc}
\end{equation}

\subsection{Pooling Graph Operation}\label{subsec:PGO}

The pooling stage of the graph is done using the Voxel downsample algorithm~\cite{open3D}. It consists of creating voxels of resolution $r_p^l$ over the point cloud and replacing all points inside the voxel with their centroid. $r_p^l$ is the radius for each pooling layer of the network and it gets increased at each pooling layer to reduce the total number of nodes in the graph. 

The feature of the new point is the average or the maximum (depends on the kind of pooling done) of the features of the points inside the voxel. After the pooling operation is done, the graph is reconstructed from the downsampled 3D point cloud.

\subsection{Architecture}\label{subsec:architecture}
The final proposed architecture is based on the well-known ResNet-18. The details of the proposal can be seen in Table~\ref{tab:architecture}. It is composed of the following basic modules:
    
\begin{itemize}
\item \textbf{Graph Init:} It creates the initial graph from the input 3D point cloud. It can connect nodes using a {kNN} approach or by {radius} proximity. As explained in section~\ref{subsec:AGC}, radius policy guarantees that each node has the same receptive field. For this reason in this work, {radius} policy will be applied.
Node features will start with a value of $1$ as the application of this work is to obtain the geometric characteristics of the scene. Edge attributes include the geometric relationship between the two connected nodes. Specifically, edge attributes in this work will be the Spherical offsets between nodes.

\item \textbf{AGC layers:} Attention graph convolution layer presented in Section~\ref{subsec:AGC} followed by an activation layer. 

\item \textbf{RAGC Block:} It consists of the stack of two AGC layers with their correspondent activation layers, as it is depicted in Figure~\ref{fig:ragc}. 

\item \textbf{Max Pooling layers:} It consists of a downsampling of the graph by using larger values of $r_p^l$ at every pooling as described in section~\ref{subsec:PGO}. Values of $r_p^l$ are selected to approximately reduce the number of nodes in the graph by a factor of two. Table~\ref{tab:radius_used} shows the radius selected for the two datasets used in this work. 

\item \textbf{Average Pooling Layer:} Similar to the previous layer but with averaging of features instead of using the maximum. The main purpose of averaging is to group all the features obtained by previous layers before the final FC layers.

\item{\textbf{FC layers}}: The last part of the proposed architecture is composed of two FC layers. These layers are in charge of classifying the 3D point cloud using the descriptor obtained by the previous stages. Both layers are followed by an activation function. Notice that the output size of the last FC layer depends on the number of classes that are intended to classify.
\end{itemize}

All activation functions chosen for this architecture are {ReLUs}, except for the last FC layer that is followed by a Softmax activation. In the training phase, a Batch Normalization block~\cite{BatchNormalization} is added between all blocks (except pooling blocks) to reduce the covariance shift. Finally, a dropout module with $0.2$ probability is incorporated between the last two FC layers to avoid overfitting. 

\begin{table}[ht]
\begin{center}
\begin{tabular}{|c|c|c|}
\hline 
\textbf{Layer Name}  & \textbf{N. output Features}  & \textbf{N. blocks}\\ 
\hline \hline
Graph Init           & - & 1\\
ACG                  & 16  & 1\\
Max Pooling          & - & 1\\
RACG Block           & 16 & 2\\
Max Pooling          & - & 1\\
RACG Block           & 32       & 2\\
Max Pooling          & -          & 1\\
RACG Block           & 64        & 2\\
Max Pooling          & -         & 1 \\
RACG Block           & 128        & 2\\
Max Pooling          & -           & 1\\
Global Average       & -           & 1\\
FC               & 128        & 1\\
FC               & Problem Dependent & 1 \\
\hline
\end{tabular}
\end{center}
\caption{Final architecture  for the proposed Residual Attention Graph Convolutional Network. }
\label{tab:architecture}
\end{table}

\section{Experiments}\label{sec:exp}
\subsection{Datasets}\label{subsec:datasets}

\begin{figure*}[!htb]
    \centering
    \begin{tabular}{cc}
    {\small (a) NYU Depth v1 Train Distribution} & {\small (b) NYU Depth v1 Test Distribution} \\
    \includegraphics[width=0.4\textwidth]{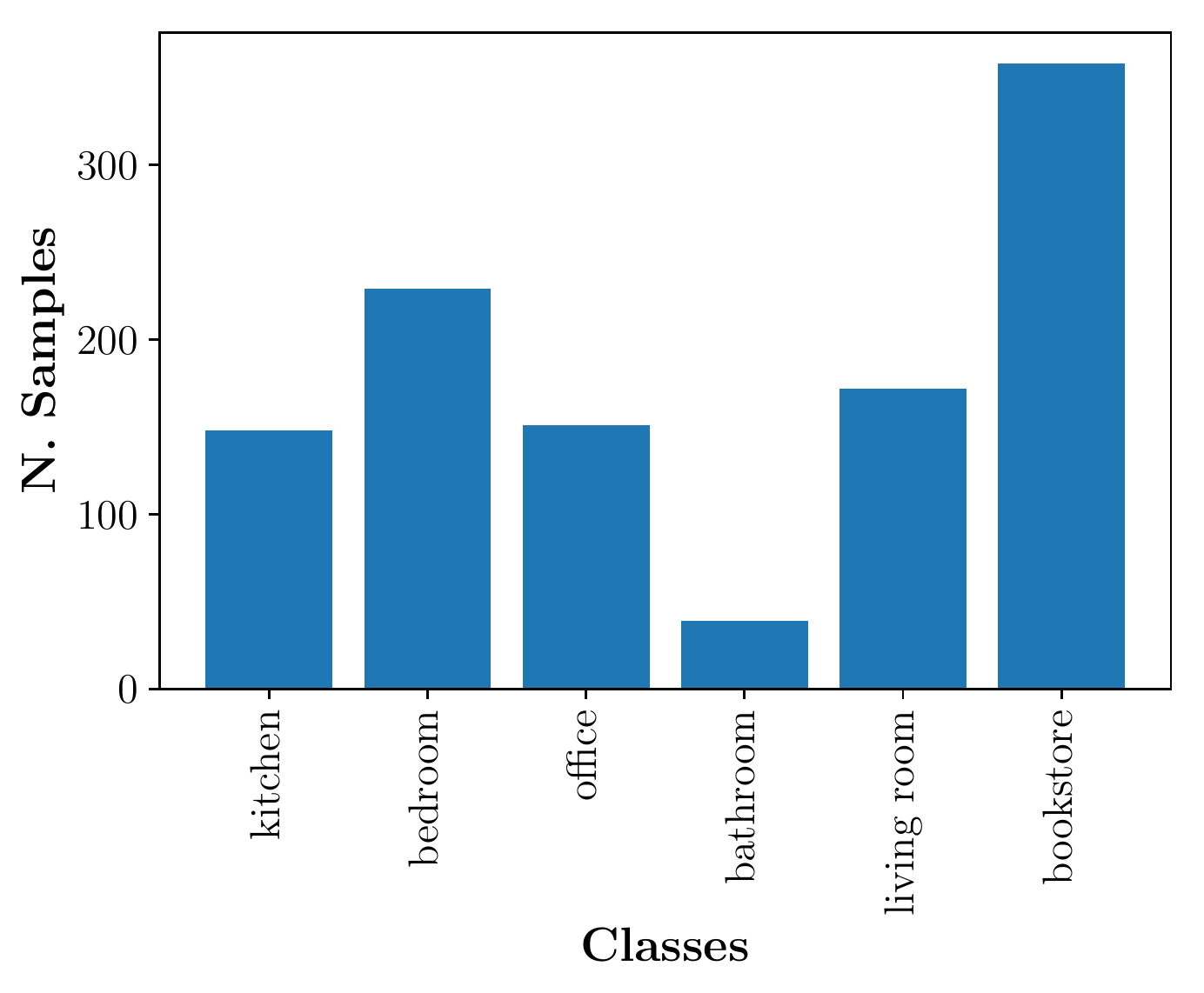} &
    \includegraphics[width=0.4\textwidth]{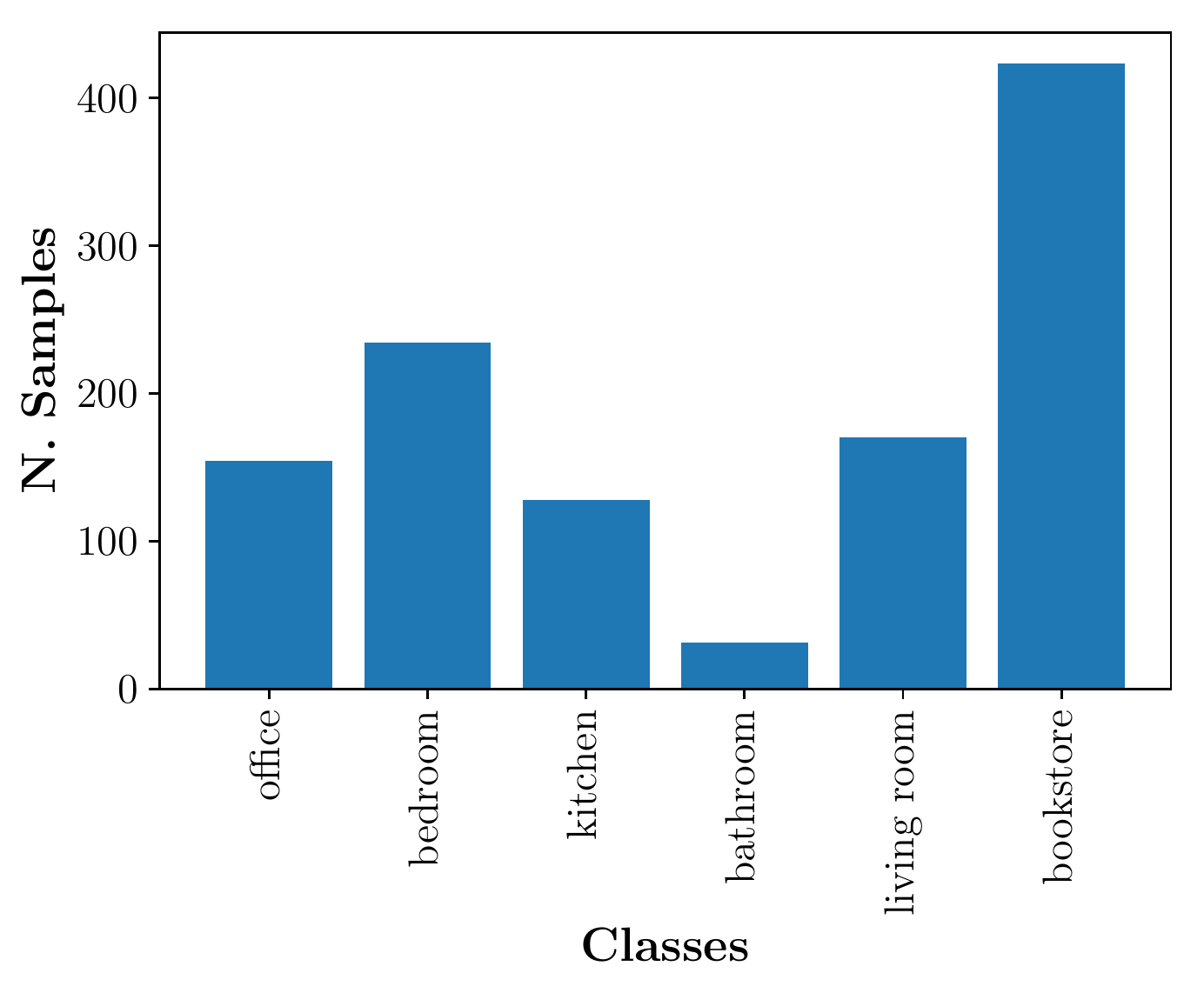} \\
    {\small (c) SUNRGBD Train Distribution} & {\small (d) SUNRGBD Test Distribution} \\
    \includegraphics[width=0.4\textwidth]{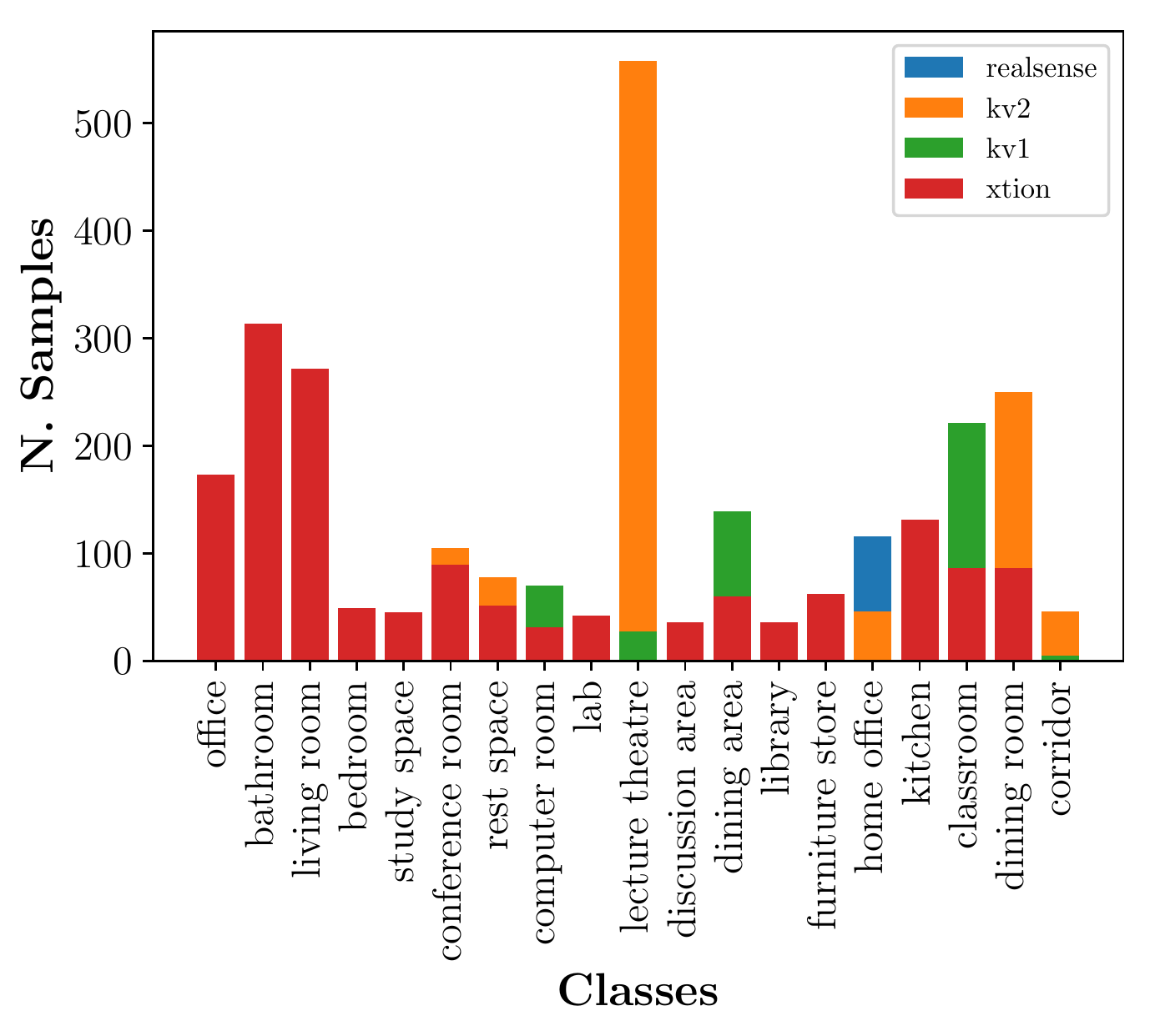} &
    \includegraphics[width=0.4\textwidth]{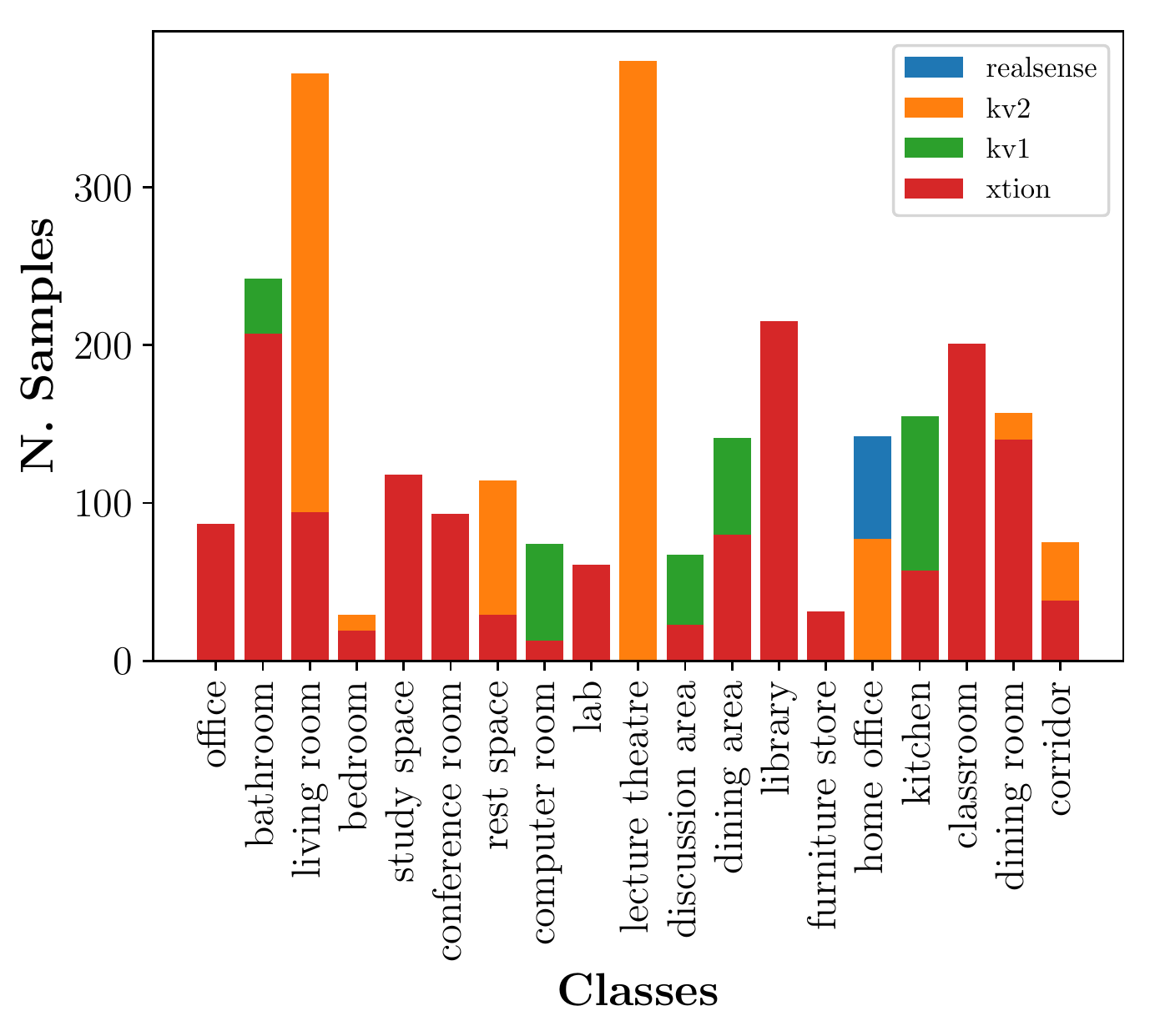} \\
    \end{tabular}
    \caption{Samples distribution of train and test splits on NYU Depth v1  dataset (first row) and SUN RGBD dataset (second row).}
    \label{fig:dataset_distribution}
\end{figure*}

Two datasets are used to evaluate the performance of the proposed method.

\textbf{NYU Depth v1} dataset created by \emph{Silberman et al.}~\cite{NYUv1} from New York University. It is composed of 2284 captures from 7 different scenes. For the indoor scene classification challenge, the \emph{cafe} scene is discarded due to the lack of variability (as recommended by the authors). The dataset is divided using the official split into 1097 captures for training and 1140 for test. Notice, that after discarding the \emph{cafe} scenario 2237 captures remain in the dataset. Figure~\ref{fig:dataset_distribution} depicts the class distribution. Hyperparameter search and ablation studies are performed in this dataset.

\textbf{SUN-RGBD} dataset created by \emph{Song et al.}~\cite{SUNRGBD} from Princeton University. It is composed of 10335 RGB-D samples. It was captured from different sensors including Asus Xtion, RealSense, Kinect v1 and Kinect v2. Following the approach proposed in the paper, classes with less than 80 samples are discarded. After that, 9504 samples remain with 19 different scenes. These samples were divided in 4845 for training and 4659 for test using the official split. Figure~\ref{fig:dataset_distribution} depicts the distribution between classes and sensors. Moreover, it can be observed that the dataset is highly unbalanced which will affect results as explained later in the section. Furthermore, there are huge differences between the number of captures done by Kinect v2 and RealSense sensors. 

For both datasets the following pre-processing is applied: a) Center crop is applied. New RGB-D resolution is $560 \times 400$ pixels. b) Downsampling by a factor of $\times 8$ so finally $3500$ RGB-D pixels are used. c) Project the RGB-D pixels to a 3D space as explained in Section~\ref{subsec:gc} to compute the initial 3D point cloud. Colour is discarded.

\subsection{Comparison with State-of-the-art}

\begin{table}[ht]
\begin{center}
\begin{tabular}{|c|c|c|}
\hline 
\multirow{2}{*}{\textbf{Layer}}  & \textbf{Graph Radius}  & \textbf{Pooling Radius}\\ 
    & $r_g$ & $r_p$ \\
\hline \hline
Graph Init & 0.1 & - \\
1. Max Pooling & 0.15 & 0.1 \\
2. Max Pooling & 0.25 & 0.15 \\
3. Max Pooling & 0.35 & 0.25 \\
4. Max Pooling & 0.55 & 0.35 \\
5. Max Pooling & 0.55 & 0.55 \\
Global Avg & - & - \\
\hline
\end{tabular}
\end{center}
\caption{Radius in meters used on the proposed network for graph construction and pooling in both architectures.}
\label{tab:radius_used}
\end{table}

\begin{table}[ht]
\begin{center}
\begin{tabular}{|c|c|}
\hline 
\textbf{Method}                                                     & \textbf{Accuracy (\%)} \\ 
\hline \hline
CNN-RNN~\cite{CRNN}                                                 & 65.2 \\
RICA~\cite{ICA}                                                     & 64.7  \\
Places2-CNNs~\cite{placesCNN}                                       & 66.9 \\
Hybrid-CNNs~\cite{placesCNN}                                        & 68.2 \\
LM-CNN~\cite{rgbdSceneCategorizationMultimodal}                     & 67.8 \\
Geometric Residual DEC~\cite{dgcnn}                                 & 58.4 \\
Geometric Residual FeaStNet~\cite{FeaStNetCVPR2018}                 & 68.6 \\
\textbf{Geometric RAGC Network}                                     & \textbf{74.5} \\
\hline
\end{tabular}
\end{center}
\caption{Comparison results of the method proposed and other published methods on NYU Depth v1 dataset obtained from article~\cite{rgbdSceneCategorizationMultimodal}. Dynamic-Edge-Convolution (DEC) and FeaStNet results are obtained after the adaptation to the scene classification problem.}
\label{tab:nyu_results}
\end{table}

\begin{table}[ht]
\begin{center}
\begin{tabular}{|c|c|}
\hline 
\textbf{Method}                                                                     & \textbf{Accuracy (\%)} \\ 
\hline \hline
CNN-RNN~\cite{CRNN}                                                                 & 26.1 \\
Places-CNN+RBF kernel SVM~\cite{placesCNN}                                          & 27.7 \\
Places2-CNNs+softmax+Alexnet~\cite{placesCNN}                                       & 32.1 \\
Places2-CNNs+softmax+VGG~\cite{placesCNN,VGG}                                       & 34.7 \\
LM-CNN~\cite{rgbdSceneCategorizationMultimodal}                                     & 34.6 \\
Geometric Residual DEC~\cite{dgcnn}                                                 & 19.0 \\
Geometric Residual FeaStNet~\cite{FeaStNetCVPR2018}                                 & 22.7\\
\textbf{Geometric RAGC Network}                                                     & \textbf{42.1} \\
\hline
\end{tabular}
\end{center}
\caption{Comparison results of the method proposed and other published methods on SUNRGBD dataset are obtained from article~\cite{rgbdSceneCategorizationMultimodal}. Dynamic-Edge-Convolution (DEC) and FeaStNet results are obtained after the adaptation to the scene classification problem.}
\label{tab:sun_results}
\end{table}

In this section, the proposed method is compared with state-of-the-art methods that use geometric information obtained from the depth channel to perform scene classification. Moreover, \emph{FeaStNet}~\cite{FeaStNetCVPR2018} and \emph{Dynamic Edge Convolution (DEC)}~\cite{dgcnn} are used over the NYU and SUN-RGBD datasets to demonstrate that the \emph{AGC} outperforms the recent state-of-the-art. \emph{AGC} exploits the local geometric context intrinsic in the 3D space. In addition, the proposed method is an end-to-end learning method opposed to LM-CNN~\cite{rgbdSceneCategorizationMultimodal} that uses a region proposal algorithm (which needs to be trained separately). Methods CNN-RNN~\cite{CRNN} and RICA~\cite{ICA} were not initially designed for geometric scene classification but are included as a reference as in~\cite{rgbdSceneCategorizationMultimodal}. 

For the proposed method, an implementation on {PyTorch}~\cite{pytorch} using Pytorch Geometric~\cite{pygeometric} is publicly available at \textbf{https://imatge-upc.github.io/ragc/}. 

For both datasets the proposed method will have the following configuration: a) {Radius} policy to generate the edges. b) Edge attributes are Spherical offsets between connected nodes giving edge attributes a total dimension of 3. c) Node features are initially set to $1$ so only geometric (no colour) information is fed into the proposed network. d) Dynamic Filter Networks used in both AGC layer and RAGC blocks to generate convolution weights from edge attributes have the following configuration: FC(16)-ReLu-FC(32)-ReLu-FC($d^l \cdot d^{l-1}$), where $d^l$ is the number of node features of layer $l$. e) Table~\ref{tab:radius_used} shows the radius used in the graph construction steps for each pooling layer. f) The proposed network is trained during 200 epochs using ADAM~\cite{ADAM} optimizer with a learning rate of: $1e^{-3}$, weigh decay of: $5e^{-4}$ and deltas in range $(0.9,0.999)$. g) Online data augmentation is performed to avoid overfitting. The following augmentation is performed during the training phase: 1) Rotation over the vertical axis randomly between $(0,2\pi)$. 2) Mirror over horizontal axes randomly with a probability of $0.5$. 3) Random removal of points in the input 3D point cloud with a probability of $0.2$.

All parameters of the network are initialized randomly in both datasets experiments. For all experiments, the maximum accuracy for five executions using random initialization seeds are reported. In the case of the experiments using \emph{FeaStNet} and \emph{DEC}, the same proposed architecture is used, where the differences are: a) AGC operation is replaced with the correspondent new graph convolutional operation, this replacement is also done inside the residual block. b) In both \emph{FeaStNet} and \emph{DEC}, the spatial coordinates are used as node features, due to the fact that both operations ignore the attributes of the edges. Based on the results reported in the respective papers and several tests, the corresponding hyperparameters are chosen. In the case of \emph{DEC} the best value of k-neighbours found is 9. For \emph{FeaStNet} the best value of \emph{M} that indicates the number of weight filters is 8 and \emph{Max} is used as aggregation stage.

\begin{figure*}[!htb]
    \centering
    \begin{tabular}{cc}
    {\small (a) NYU Depth v1 Confusion Matrix} & {\small (b) SUN-RGBD Confusion Matrix} \\
    \includegraphics[width=0.4\textwidth]{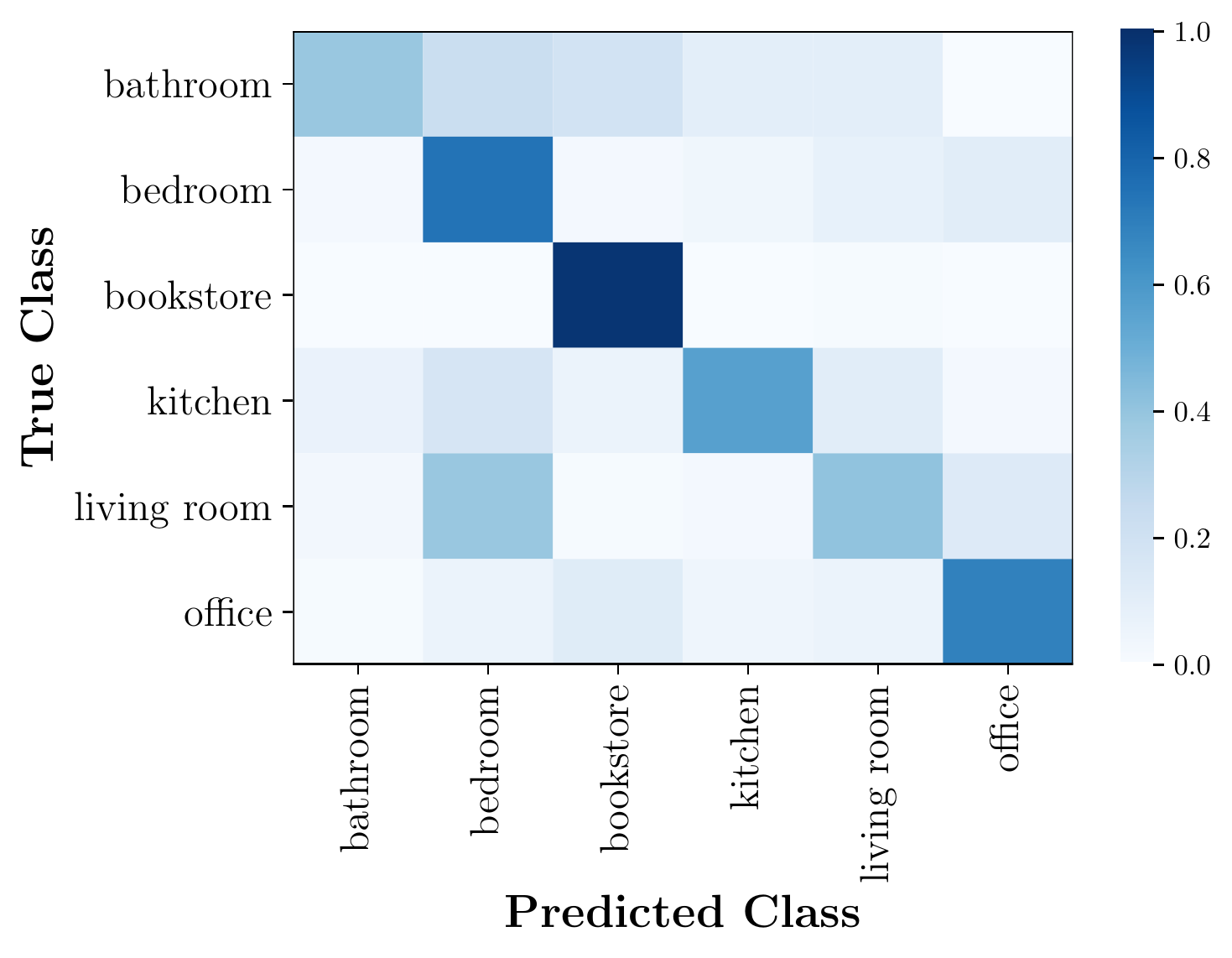} &
    \includegraphics[width=0.4\textwidth]{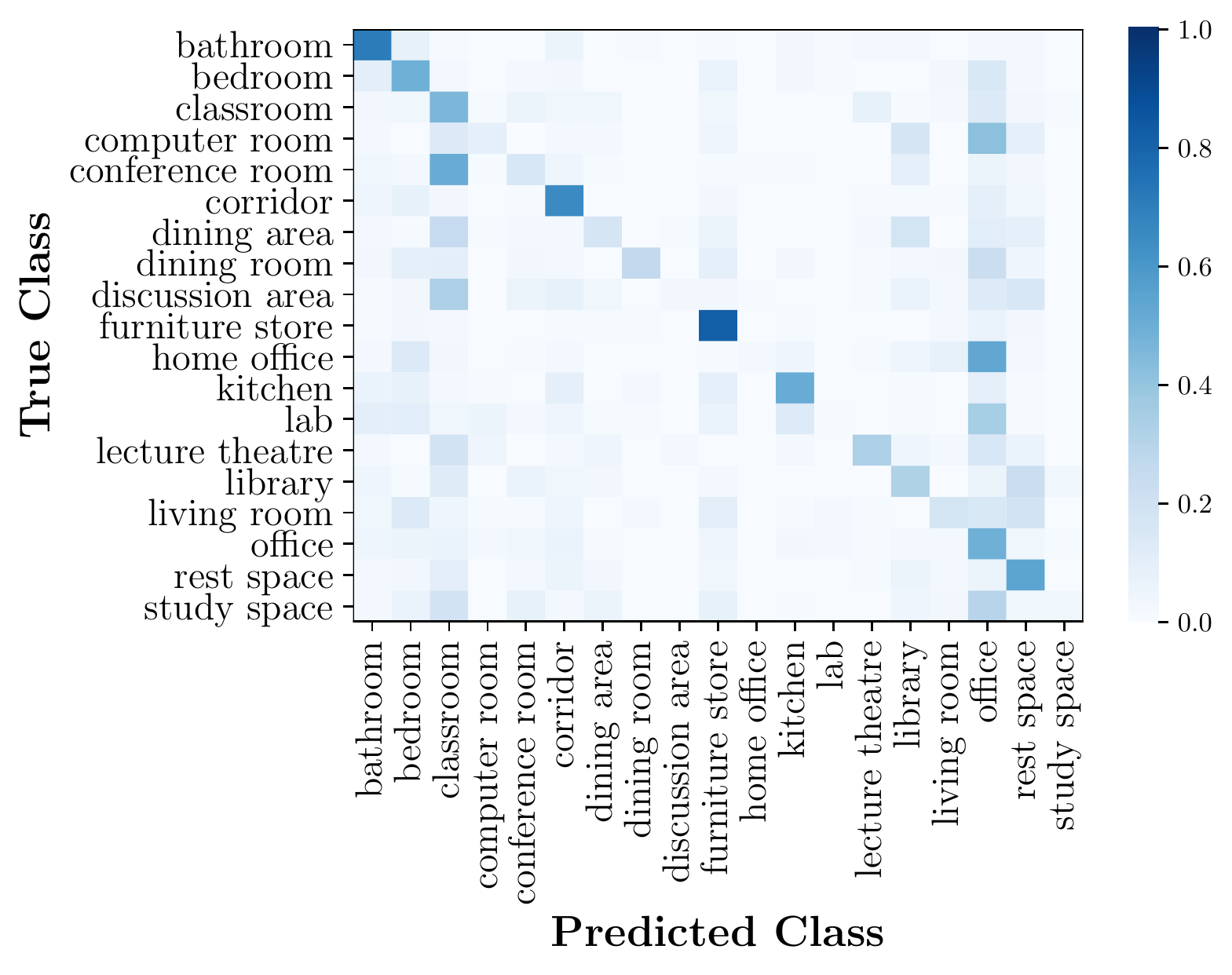} \\
    \end{tabular}
    \caption{Confusion matrixes of the proposed method on NYU Depth v1 dataset (left) and SUN RGBD dataset (right). The labels on the vertical axis express the true classes and the labels on the horizontal axis denote the predicted classes.}
    \label{fig:conf_matrix}
\end{figure*}

As can be observed in Tables~\ref{tab:nyu_results} and~\ref{tab:sun_results} the proposed method outperforms current state-of-the-art. In the case of NYU Depth v1 dataset, the proposed method exceeds by $6\%$ the accuracy of previous scene classification approaches. Moreover, \emph{RAGC} surpasses \emph{DEC} and \emph{FeaStNet} in this specific problem. Both methods ignore edge attributes that limit the representation of the relationship of the nodes with their neighbourhood. Furthermore, the formulation of \emph{DEC} limits the operation in the use of k-neighbours, this policy does not guarantee to have the same receptive field in 3D point clouds that leads to having poor results in complex point clouds. In the case of the more challenging SUN-RGBD dataset, the proposed method still improves other scene classification state-of-the-art methods with an increase of $7.4\%$ of the accuracy and surpasses \emph{DEC} and \emph{FeaStNet}. Demonstrating that the proposed method is able to obtain better geometric representation of the scene than the current state-of-the-art methodologies.
Figure~\ref{fig:conf_matrix} shows the confusion matrix for each dataset. The diagonal elements represent the recognition accuracy for each category. In the case of the NYU Depth v1 dataset, the confusion matrix shows that the classes have more or less the same percentage of accuracy except for \emph{bookstore}, which presents higher accuracy. The confusion matrix for the SUN-RGBD shows problems to classify some of the classes which might be due to the unbalance of the original dataset as explained in Section~\ref{subsec:datasets}.

\subsection{Ablation Studies}\label{subsec:ablation_studies}
In the previous section, the best configuration of the proposed method is compared against the state-of-the-art. In this section, different configurations will be explored using NYU Depth V1 dataset. Specifically, different edge attributes, configurations of the Dynamic Filter Network and the influence of the residual block will be studied using the proposed architecture as baseline.

\textbf{Edge Generation Policies.} The edge generation policies studied will be {kNN} and radius policies previously explained. For the purpose of this experiment the attributes of the edge will be fixed using Spherical offsets, the configuration of the Dynamic Filter Network will be fixed as FC(16)-ReLu-FC(32)-ReLu-FC($d_l \cdot d_{l-1})$ and residual blocks will be used. After several experiments it was found that the best value of {k} was 9. Table~\ref{tab:edge_policies} shows the results of these experiments using the best k found. As can be observed, radius policy surpasses the accuracy obtained with {kNN}. 

\begin{table}[ht]
\begin{center}
    \begin{tabular}{|c|c|}
    \hline 
    \textbf{Policy}                                              & \textbf{Accuracy(\%)} \\
    \hline \hline
   \textbf{Radius}                                             & \textbf{74.5}\\
    {kNN}                                                        & 70.9 \\
    \hline
    \end{tabular}
\end{center}
    \caption{Comparison of edge generation policies.}
    \label{tab:edge_policies}
\end{table}

\textbf{Edge Attributes.} The edge attributes studied will be: Cartesian offsets, Spherical offsets and the concatenation of Cartesian and Spherical offsets. For the purpose of this experiment the radius policy will be used, the configuration of the Dynamic Filter Network will be fixed as FC(16)-ReLu-FC(32)-ReLu-FC($d_l \cdot d_{l-1})$ and residual blocks will be used. Table~\ref{tab:edge_features_study} shows the results of these experiments. As can be observed, when Spherical offsets are used the results are better than Cartesian offsets or the combination of both offsets. 

\begin{table}[ht]
\begin{center}
    \begin{tabular}{|c|c|}
    \hline 
    \textbf{Edge Feature}                       & \textbf{Accuracy(\%)} \\
    \hline \hline
    Cartesian Offset                            & 72.3\\
    \textbf{Spherical Offset}                   &\textbf{ 74.5} \\
    Cartesian and Spherical Offset         & 73.2 \\
    \hline
    \end{tabular}
\end{center}
    \caption{Comparison of different edges attributes of the proposed method.}
    \label{tab:edge_features_study}
\end{table}

\begin{table}[ht]
\begin{center}
    \begin{tabular}{|c|c|}
    \hline 
    \textbf{Configuration}                                              & \textbf{Accuracy(\%)} \\
    \hline \hline
    FC($d_ld_{l-1}$)                                              & 70\\
    FC(16)-FC($d_ld_{l-1}$)                                        & 71.6 \\
    FC(32)-FC($d_ld_{l-1}$)                                        & 74.1 \\
    \textbf{FC(16)-FC(32)-FC($\mathbf{d_ld_{l-1}}$) }         & \textbf{74.5} \\
    \hline
    \end{tabular}
\end{center}
    \caption{Comparison of different configurations of the Dynamic Filter Network. Note that between two consecutive FCs there is an activation function.}
    \label{tab:filter_network_study}
\end{table}

\textbf{Dynamic Filter Network.} To study the influence of the Dynamic Filter Network architecture the attributes of the edge will be fixed using the Spherical offsets. Moreover residual blocks and radius policy will be used. In Table~\ref{tab:filter_network_study} the different cases analyzed and their results are shown. As can be observed, there is a slight difference between the two and three FCs layer configuration. This means that it is possible to compress the network without a significant loss of performance. Furthermore, it is possible to compress the network using a single FC configuration with a loss of $4.5\%$ accuracy and still achieve a performance which is better than the current state-of-the-art.

\textbf{Residual Graph Network.} To study the influence of residual blocks in the architecture, the attributes of the edge will be fixed using Spherical offsets, the configuration of the Dynamic Filter Network will be fixed as FC(16)-ReLu-FC(32)-ReLu-FC($d_l \cdot d_{l-1})$ and radius policy will be used. Table~\ref{tab:residual_study} shows the results of these experiments. As can be observed, the residual blocks help to increase the accuracy in $1.9\%$.

\begin{table}[ht]
\begin{center}
    \begin{tabular}{|c|c|}
    \hline 
    \textbf{Configuration}                                              & \textbf{Accuracy(\%)} \\
    \hline \hline
   \textbf{ Residual}                                             & \textbf{74.5}\\
    Plain                                        & 72.4 \\
    \hline
    \end{tabular}
\end{center}
    \caption{Comparison of plain and residual configurations using same number of layers.}
    \label{tab:residual_study}
\end{table}

\section{Conclusions}\label{sec:conc}
This work presents a novel Residual Attention Graph Convolutional Network that outperforms the current state-of-the-art in Geometric 3D Scene Classification. This proposal exploits the geometric context intrinsic in 3D space that helps the network to learn the geometric relations between points in a 3D point cloud. This is possible due to the fact that the proposed network is able to focus on the important relationships between points in the 3D point cloud. Different depth sensors can be used as node features are not required. The proposed network extends current Graph Convolutional Networks to a deep architecture in a residual learning framework similar to ResNet in standard convolutional networks. In the future, new attributes for the edge will be investigated to improve the generation of convolutional weights by the Dynamic Filter Network. Furthermore, the attention stage will be improved studying new projection functions that take into account edge features in successive layers. 

\section{Acknowledgments}
This research was supported by Secretary of Universities and Research of the Generalitat de Catalunya and the European Social Fund via a PhD grant (FI2019) in the framework of project TEC2016-75976-R, financed by the Ministerio de Econom\'ia, Industria y Competitividad and the European Regional Development Fund (ERDF).


{\small
\bibliographystyle{ieee}
\bibliography{bibliography}

\begin{thebibliography}{10}\itemsep=-1pt

\bibitem{machineTranslationAttention}
D.~Bahdanau, K.~Cho, and Y.~Bengio.
\newblock {Neural Machine Translation by Jointly Learning to Align and
  Translate}.
\newblock {\em International Conference on Learning Representations (ICLR},
  2015.

\bibitem{SnapNet}
A.~Boulch, B.~L. Saux, and N.~Audebert.
\newblock {Unstructured Point Cloud Semantic Labeling Using Deep Segmentation
  Networks}.
\newblock In I.~Pratikakis, F.~Dupont, and M.~Ovsjanikov, editors, {\em
  Eurographics Workshop on 3D Object Retrieval}. The Eurographics Association,
  2017.

\bibitem{GatedGraphConv}
X.~Bresson and T.~Laurent.
\newblock An experimental study of neural networks for variable graphs.
\newblock {\em ICLR 2018 Workshop}, 2018.

\bibitem{SIFT}
M.~{Brown} and S.~{Süsstrunk}.
\newblock Multi-spectral sift for scene category recognition.
\newblock In {\em CVPR 2011}, pages 177--184, June 2011.

\bibitem{BrunaSpectralNetworks}
J.~Bruna, W.~Zaremba, A.~Szlam, and Y.~LeCun.
\newblock {Spectral Networks and Locally Connected Networks on Graphs}.
\newblock {\em CoRR}, abs/1312.6, 2013.

\bibitem{rgbdSceneCategorizationMultimodal}
Z.~Cai and L.~Shao.
\newblock Rgb-d scene classification via multi-modal feature learning.
\newblock {\em Cognitive Computation}, Aug 2018.

\bibitem{remoteSensingSceneClassification}
G.~{Cheng}, Z.~{Li}, X.~{Yao}, L.~{Guo}, and Z.~{Wei}.
\newblock Remote sensing image scene classification using bag of convolutional
  features.
\newblock {\em IEEE Geoscience and Remote Sensing Letters}, 14(10):1735--1739,
  Oct 2017.

\bibitem{ScanNet}
A.~Dai, A.~X. Chang, M.~Savva, M.~Halber, T.~Funkhouser, and M.~Nie{\ss}ner.
\newblock {ScanNet: Richly-annotated 3D reconstructions of indoor scenes}.
\newblock In {\em Proceedings - 30th IEEE Conference on Computer Vision and
  Pattern Recognition, CVPR 2017}, volume 2017-Janua, pages 2432--2443. IEEE, 7
  2017.

\bibitem{dai3DMV}
A.~Dai and M.~Nie{\ss}ner.
\newblock {3DMV: Joint 3D-Multi-view Prediction for 3D Semantic Scene
  Segmentation}.
\newblock In V.~Ferrari, M.~Hebert, C.~Sminchisescu, and Y.~Weiss, editors,
  {\em Computer Vision -- ECCV 2018}, pages 458--474, Cham, 2018. Springer
  International Publishing.

\bibitem{DefferrardCNNGSpectral}
M.~Defferrard, X.~Bresson, and P.~Vandergheynst.
\newblock {Convolutional Neural Networks on Graphs with Fast Localized Spectral
  Filtering}.
\newblock {\em NIPS}, 2016.

\bibitem{DuvenaudCNG}
D.~Duvenaud, D.~Maclaurin, J.~Aguilera-Iparraguirre, R.~G\'{o}mez-Bombarelli,
  T.~Hirzel, A.~Aspuru-Guzik, and R.~P. Adams.
\newblock Convolutional networks on graphs for learning molecular fingerprints.
\newblock In {\em Proceedings of the 28th International Conference on Neural
  Information Processing Systems - Volume 2}, NIPS'15, pages 2224--2232,
  Cambridge, MA, USA, 2015. MIT Press.

\bibitem{pygeometric}
M.~Fey and J.~E. Lenssen.
\newblock Fast graph representation learning with {PyTorch Geometric}.
\newblock In {\em ICLR Workshop on Representation Learning on Graphs and
  Manifolds}, 2019.

\bibitem{semantincSceneRecognition}
M.~George, M.~Dixit, G.~Zogg, and N.~Vasconcelos.
\newblock Semantic clustering for robust fine-grained scene recognition.
\newblock In B.~Leibe, J.~Matas, N.~Sebe, and M.~Welling, editors, {\em
  Computer Vision -- ECCV 2016}, pages 783--798, Cham, 2016. Springer
  International Publishing.

\bibitem{SnapNetR}
J.~Guerry, A.~Boulch, B.~L. Saux, J.~Moras, A.~Plyer, and D.~Filliat.
\newblock {SnapNet-R: Consistent 3D Multi-view Semantic Labeling for Robotics}.
\newblock In {\em 2017 IEEE International Conference on Computer Vision
  Workshops (ICCVW)}, pages 669--678. IEEE, 10 2017.

\bibitem{ResNet}
K.~He, X.~Zhang, S.~Ren, and J.~Sun.
\newblock {Deep Residual Learning for Image Recognition}.
\newblock In {\em 2016 IEEE Conference on Computer Vision and Pattern
  Recognition (CVPR)}, 2016.

\bibitem{BatchNormalization}
S.~Ioffe and C.~Szegedy.
\newblock Batch normalization: Accelerating deep network training by reducing
  internal covariate shift.
\newblock In F.~Bach and D.~Blei, editors, {\em Proceedings of the 32nd
  International Conference on Machine Learning}, volume~37 of {\em Proceedings
  of Machine Learning Research}, pages 448--456, Lille, France, 07--09 Jul
  2015. PMLR.

\bibitem{Dynamicfilter}
X.~Jia, B.~De~Brabandere, T.~Tuytelaars, and L.~V. Gool.
\newblock {Dynamic Filter Networks}.
\newblock In D.~D. Lee, M.~Sugiyama, U.~V. Luxburg, I.~Guyon, and R.~Garnett,
  editors, {\em Advances in Neural Information Processing Systems 29}, pages
  667--675. Curran Associates, Inc., 2016.

\bibitem{ADAM}
D.~Kingma and J.~Ba.
\newblock Adam: A method for stochastic optimization.
\newblock {\em International Conference on Learning Representations}, 12 2014.

\bibitem{kipf2017semi}
T.~N. Kipf and M.~Welling.
\newblock {Semi-Supervised Classification with Graph Convolutional Networks}.
\newblock In {\em International Conference on Learning Representations (ICLR)},
  2017.

\bibitem{ICA}
Q.~V. Le, A.~Karpenko, J.~Ngiam, and A.~Y. Ng.
\newblock Ica with reconstruction cost for efficient overcomplete feature
  learning.
\newblock In J.~Shawe-Taylor, R.~S. Zemel, P.~L. Bartlett, F.~Pereira, and
  K.~Q. Weinberger, editors, {\em Advances in Neural Information Processing
  Systems 24}, pages 1017--1025. Curran Associates, Inc., 2011.

\bibitem{FCNSemanticSegmentation}
J.~Long, E.~Shelhamer, and T.~Darrell.
\newblock {Fully convolutional networks for semantic segmentation}.
\newblock In {\em 2015 IEEE Conference on Computer Vision and Pattern
  Recognition (CVPR)}, 2015.

\bibitem{VoxNet}
D.~Maturana and S.~Scherer.
\newblock {VoxNet: A 3D Convolutional Neural Network for real-time object
  recognition}.
\newblock In {\em IEEE International Conference on Intelligent Robots and
  Systems}, volume 2015-Decem, pages 922--928, 2015.

\bibitem{MoNet}
F.~Monti, D.~Boscaini, J.~Masci, E.~Rodol{\`{a}}, J.~Svoboda, and M.~M.
  Bronstein.
\newblock {Geometric deep learning on graphs and manifolds using mixture model
  CNNs}.
\newblock In {\em CVPR}, 2017.

\bibitem{pytorch}
A.~Paszke, S.~Gross, S.~Chintala, G.~Chanan, E.~Yang, Z.~DeVito, Z.~Lin,
  A.~Desmaison, L.~Antiga, and A.~Lerer.
\newblock Automatic differentiation in pytorch.
\newblock In {\em NIPS-W}, 2017.

\bibitem{VolumetricMultiViewCNN}
C.~R. Qi, H.~Su, M.~Niessner, A.~Dai, M.~Yan, and L.~J. Guibas.
\newblock {Volumetric and Multi-View CNNs for Object Classification on 3D
  Data}.
\newblock {\em 2016 IEEE Conference on Computer Vision and Pattern Recognition
  (CVPR)}, pages 5648--5656, 6 2016.

\bibitem{Qui3DGNNSS}
X.~Qi, R.~Liao, J.~Jia, S.~Fidler, and R.~Urtasun.
\newblock {3D Graph Neural Networks for RGBD Semantic Segmentation}.
\newblock In {\em 2017 IEEE International Conference on Computer Vision
  (ICCV)}, pages 5209--5218. IEEE, 10 2017.

\bibitem{instanceSegmentationRecurrentAttention}
M.~{Ren} and R.~S. {Zemel}.
\newblock End-to-end instance segmentation with recurrent attention.
\newblock In {\em 2017 IEEE Conference on Computer Vision and Pattern
  Recognition (CVPR)}, pages 293--301, July 2017.

\bibitem{ScarselliGraphNeuralNetwork}
F.~Scarselli, M.~Gori, {Ah Chung Tsoi}, M.~Hagenbuchner, and G.~Monfardini.
\newblock {The Graph Neural Network Model}.
\newblock {\em IEEE Transactions on Neural Networks}, 20(1):61--80, 1 2009.

\bibitem{NYUv1}
N.~Silberman and R.~Fergus.
\newblock Indoor scene segmentation using a structured light sensor.
\newblock In {\em Proceedings of the International Conference on Computer
  Vision - Workshop on 3D Representation and Recognition}, 2011.

\bibitem{SimonovskyECC}
M.~Simonovsky and N.~Komodakis.
\newblock {Dynamic edge-conditioned filters in convolutional neural networks on
  graphs}.
\newblock {\em Proceedings - 30th IEEE Conference on Computer Vision and
  Pattern Recognition, CVPR 2017}, 2017-Janua:29--38, 2017.

\bibitem{VGG}
K.~Simonyan and A.~Zisserman.
\newblock {Very Deep Convolutional Networks for large-scale Image Recognition}.
\newblock {\em International Conference on Learning Representations}, 2015.

\bibitem{CRNN}
R.~Socher, B.~Huval, B.~Bhat, C.~D. Manning, and A.~Y. Ng.
\newblock Convolutional-recursive deep learning for 3d object classification.
\newblock In {\em Proceedings of the 25th International Conference on Neural
  Information Processing Systems - Volume 1}, NIPS'12, pages 656--664, USA,
  2012. Curran Associates Inc.

\bibitem{SUNRGBD}
S.~{Song}, S.~P. {Lichtenberg}, and J.~{Xiao}.
\newblock Sun rgb-d: A rgb-d scene understanding benchmark suite.
\newblock In {\em 2015 IEEE Conference on Computer Vision and Pattern
  Recognition (CVPR)}, pages 567--576, June 2015.

\bibitem{MultiViewSu3DShapeRecog}
H.~Su, S.~Maji, E.~Kalogerakis, and E.~Learned-Miller.
\newblock {Multi-view Convolutional Neural Networks for 3D Shape Recognition}.
\newblock In {\em 2015 IEEE International Conference on Computer Vision
  (ICCV)}, pages 945--953. IEEE, 12 2015.

\bibitem{peernets}
J.~Svoboda, J.~Masci, F.~Monti, M.~Bronstein, and L.~Guibas.
\newblock {PeerNets: Exploiting Peer Wisdom Against Adversarial Attacks}.
\newblock In {\em International Conference on Learning Representations}, 2019.

\bibitem{OctreeGeneratingNetworks}
M.~Tatarchenko, A.~Dosovitskiy, and T.~Brox.
\newblock {Octree Generating Networks: Efficient Convolutional Architectures
  for High-resolution 3D Outputs}.
\newblock In {\em Proceedings of the IEEE International Conference on Computer
  Vision}, volume 2017-Octob, pages 2107--2115. IEEE, 10 2017.

\bibitem{SEGCloud}
L.~Tchapmi, C.~Choy, I.~Armeni, J.~Gwak, and S.~Savarese.
\newblock {SEGCloud: Semantic segmentation of 3D point clouds}.
\newblock In {\em Proceedings - 2017 International Conference on 3D Vision, 3DV
  2017}, pages 537--547, 2017.

\bibitem{HoG}
M.~Uddin.
\newblock Scene classification using localized histogram of oriented gradients
  method.
\newblock {\em International Journal of Computer (IJC)}, 20:13--18, 01 2016.

\bibitem{GAT2018}
P.~Veli{\v{c}}kovi{\'{c}}, G.~Cucurull, A.~Casanova, A.~Romero, P.~Li{\`{o}},
  and Y.~Bengio.
\newblock {Graph Attention Networks}.
\newblock {\em International Conference on Learning Representations}, 2018.

\bibitem{FeaStNetCVPR2018}
N.~Verma, E.~Boyer, and J.~Verbeek.
\newblock {FeaStNet: Feature-Steered Graph Convolutions for 3D Shape Analysis}.
\newblock In {\em CVPR}, 2018.

\bibitem{dgcnn}
Y.~Wang, Y.~Sun, Z.~Liu, S.~E. Sarma, M.~M. Bronstein, and J.~M. Solomon.
\newblock {Dynamic Graph CNN for Learning on Point Clouds}.
\newblock {\em ACM Transactions on Graphics (TOG)}, 2019.

\bibitem{3DShapeNets}
Z.~Wu, S.~Song, A.~Khosla, F.~Yu, L.~Zhang, X.~Tang, and J.~Xiao.
\newblock {3D ShapeNets: A deep representation for volumetric shapes}.
\newblock In {\em Proceedings of the IEEE Computer Society Conference on
  Computer Vision and Pattern Recognition}, volume 07-12-June, pages
  1912--1920. IEEE, 6 2015.

\bibitem{visualAttentionNeuralImage}
K.~Xu, J.~L. Ba, R.~Kiros, K.~Cho, A.~Courville, R.~Salakhutdinov, R.~S. Zemel,
  and Y.~Bengio.
\newblock Show, attend and tell: Neural image caption generation with visual
  attention.
\newblock In {\em Proceedings of the 32Nd International Conference on
  International Conference on Machine Learning - Volume 37}, ICML'15, pages
  2048--2057. JMLR.org, 2015.

\bibitem{pspNet}
H.~Zhao, J.~Shi, X.~Qi, X.~Wang, and J.~Jia.
\newblock Pyramid scene parsing network.
\newblock In {\em Proceedings of IEEE Conference on Computer Vision and Pattern
  Recognition (CVPR)}, 2017.

\bibitem{placesCNN}
B.~Zhou, A.~Lapedriza, J.~Xiao, A.~Torralba, and A.~Oliva.
\newblock Learning deep features for scene recognition using places database.
\newblock In Z.~Ghahramani, M.~Welling, C.~Cortes, N.~D. Lawrence, and K.~Q.
  Weinberger, editors, {\em Advances in Neural Information Processing Systems
  27}, pages 487--495. Curran Associates, Inc., 2014.

\bibitem{open3D}
Q.-Y. Zhou, J.~Park, and V.~Koltun.
\newblock {Open3D}: {A} modern library for {3D} data processing.
\newblock {\em arXiv:1801.09847}, 2018.

\bibitem{semanticRecApp}
W.~{Zhuo}, M.~{Salzmann}, X.~{He}, and M.~{Liu}.
\newblock Indoor scene parsing with instance segmentation, semantic labeling
  and support relationship inference.
\newblock In {\em 2017 IEEE Conference on Computer Vision and Pattern
  Recognition (CVPR)}, pages 6269--6277, July 2017.

\end{thebibliography}
}

\end{document}